\documentclass[runningheads]{llncs}
\pdfoutput=1
\usepackage{cite}
\usepackage{amsmath,amssymb,amsfonts}
\usepackage{algorithmic}
\usepackage{graphicx}
\usepackage{textcomp}
\usepackage{xcolor}
\usepackage{tabularx} 
\usepackage{algorithm}
\usepackage{multirow}
\usepackage{booktabs}
\usepackage{subfigure}
\usepackage{pdfpages}
\usepackage{enumitem}
\usepackage{tcolorbox}
\usepackage[misc]{ifsym}

\usepackage{array}
\begin{document}
%
\title{Enhancing Question Answering for Enterprise Knowledge Bases using Large Language Models}

\titlerunning{Enhancing Question Answering for Enterprise Knowledge Bases using LLMs}



\author{Feihu Jiang\inst{1, 2}\thanks{This work was accomplished when Feihu Jiang was an intern in Career Science Lab, BOSS Zhipin, supervised by Chuan Qin and Hengshu Zhu.} \and
Chuan Qin\inst{1,3}\textsuperscript{(\Letter)} \and
Kaichun Yao\inst{4} \and
Chuyu Fang\inst{1}  \and \\
Fuzhen Zhuang\inst{5,6} \and
Hengshu Zhu\inst{1} \and
Hui Xiong\inst{7,8,9}\textsuperscript{(\Letter)}
}
%
%
\institute{Career Science Lab, BOSS Zhipin, Beijing, China\\
\email{fangchuyu@kanzhun.com}
\and
University of Science and Technology of China\\
\email{jiangfeihu@mail.ustc.edu.cn}
\and  PBC School of Finance, Tsinghua University, Beijing, China
\and
Institute of Software, Chinese Academy of Sciences\\
\and Institute of Artificial Intelligence, Beihang University,
\and Zhongguancun Laboratory\\
\email{zhuangfuzhen@buaa.edu.cn}\\
\and The Thrust of Artificial Intelligence, The Hong Kong University of Science and Technology (GuangZhou)
\and The Department of Computer Science and Engineering, The Hong Kong University of Science and Technology
\and Guangzhou HKUST Fok Ying Tung Research Institute \\
\email{xionghui@ust.hk}\\
\email{yaokaichun@outlook.com,\{chuanqin0426,  zhuhengshu\}@gmail.com}
}


\maketitle

\begin{abstract}

Efficient knowledge management plays a pivotal role in augmenting both the operational efficiency and the innovative capacity of businesses and organizations. By indexing knowledge through vectorization, a variety of knowledge retrieval methods have emerged, significantly enhancing the efficacy of knowledge management systems. Recently, the rapid advancements in generative natural language processing technologies paved the way for generating precise and coherent answers after retrieving relevant documents tailored to user queries. However, for enterprise knowledge bases, assembling extensive training data from scratch for knowledge retrieval and generation is a formidable challenge due to the privacy and security policies of private data, frequently entailing substantial costs. To address the challenge above, in this paper, we propose EKRG, a novel \textbf{R}etrieval-\textbf{G}eneration framework based on large language models (LLMs), expertly designed to enable question-answering for \textbf{E}nterprise \textbf{K}nowledge bases with limited annotation costs
. Specifically, for the retrieval process, we first introduce an instruction-tuning method using an LLM to generate sufficient document-question pairs for training a knowledge retriever. This method, through carefully designed instructions, efficiently generates diverse questions for enterprise knowledge bases, encompassing both fact-oriented and solution-oriented knowledge. Additionally, we develop a relevance-aware teacher-student learning strategy to further enhance the efficiency of the training process. For the generation process, we propose a novel chain of thought (CoT) based fine-tuning method to empower the LLM-based generator to adeptly respond to user questions using retrieved documents. Finally, extensive experiments on real-world datasets have demonstrated the effectiveness of our proposed framework.

\end{abstract}
\section{Introduction}

In the present-day, within the ever-evolving and highly competitive marketing landscape, efficient knowledge management emerges as a critical factor, 
not only in enhancing operational efficiency but also in bolstering the capacity for innovation. It can facilitate a seamless flow of information, support informed decision-making, and encourage the dissemination of domain knowledge.

Knowledge-oriented question answering (QA), a key component of enterprise knowledge management, aims to provide accurate responses to user queries by leveraging a knowledge base~\cite{voorhees1999trec}. Current QA systems combine the strengths of information retrieval and natural language processing technologies, primarily following a ``retrieve-and-read'' architecture. This involves two crucial steps: 1)~The system, upon receiving a user query, retrieves relevant documents from a knowledge base; 2) It then employs a reader model to distill and construct an answer from the content within these documents.

Traditional approaches utilize sparse retrievers, such as BM25~\cite{robertson1995okapi}, in the retrieval phase to link questions and documents via lexical overlap. However, a notable limitation is their tendency to overlook semantically pertinent documents that exhibit minimal lexical similarity to the question. Dense retrievers address this challenge by transforming questions and documents into dense vector representations. Such approaches enable the clustering of synonyms and paraphrase into similar vectors, accomplished through extensive fine-tuning on a vast collection of annotated question-document pairs~\cite{ren2022thorough}. After that, reader models utilize specialized extractors to identify answer spans within retrieved documents that have a high probability of addressing the user's question.
Recently, generative natural language processing, moving beyond mere extraction of answer spans, has significantly revolutionized the presentation of information in response to user queries. For instance, RAG~\cite{lewis2020retrieval} adopts pre-trained Seq2Seq language models to generate precise and coherent answers.

Existing QA systems, particularly those utilizing dense retrievers and generative readers, require substantial training data to precisely retrieve relevant documents from knowledge bases in response to user queries, and to subsequently produce corresponding answers that are as natural-sounding as possible. However, assembling extensive training data from scratch
for knowledge retrieval and generation poses a significant challenge for enterprise knowledge bases. Recently, the capabilities of LLMs in few-shot learning have been validated, leading researchers to explore their use for training data augmentation. For instance, \textsl{Bonifacio et al.} utilized GPT-3 to construct query-document pairs for training dense retrievers~\cite{inpars}. However, such approaches, dependent on third-party services, often fail to comply with the strict privacy and security policies that govern private data.

To this end, in this paper, we introduce EKRG, a novel retrieval-generation framework leveraging LLMs to facilitate question-answering in Enterprise Knowledge bases with limited annotation costs. Specifically, in the retrieval process, we first propose an instruction-tuning approach, using a limited set of annotated examples to align the LLMs with the task of question generation. Through carefully designed instructions, we efficiently generate diverse questions for enterprise knowledge bases, encompassing both fact-oriented and solution-oriented knowledge. Furthermore, to enhance the quality of generated document-query pairs, we employ a relevance-aware teacher-student learning strategy that iteratively updates the knowledge retriever. In the generation process, a novel CoT-based fine-tuning method is proposed to empower the LLM-based generator to adeptly respond to user questions using retrieved documents. Finally, extensive experiments conducted on real-world datasets clearly demonstrate the effectiveness of our proposed framework.

\section{Related Works}

QA is a critical task in natural language processing, enabling the extraction of factual information from extensive knowledge corpora like Wikipedia, which has been applied in a wide range of fields\cite{qin2023comprehensive}. Retriever-reader methods, currently the most effective approach in QA, comprise two key modules: a retriever, which fetches relevant documents in response to a given question, and a reader, leveraging a neural machine reading comprehension model to deduce the final answer from the retrieved documents.

\textbf{Neural Retrievers.}
Neural network-based models have enhanced data representation, thereby advancing automated learning processes\cite{qin2018enhancing,qin2020enhanced,shen2021topic,jiang2024resume}.
They have also been introduced to automatically learn query and document representations from labeled data. 
However, dense retrievers require a substantial amount of annotated data,  making their training cost-prohibitive. 
Private knowledge bases often lack annotated query-document pairs, underscoring the importance of developing dense retrieval models without relying on human-annotated data. The recent advancements in generative language models have brought new perspectives to constructing data for addressing this issue\cite{qin2023automatic,qin2022towards}.
Notably, recent efforts have demonstrated the feasibility of training dense retrievers in an annotation-free manner. For instance, models like Contriever~\cite{izacard2021unsupervised} and ICT~\cite{lee2019latent} construct pseudo query-document pairs by randomly selecting two spans from a document. InPars leverages the few-shot learning capabilities of large language models to generate reliable synthetic training data for information retrieval models using only a few supervised examples~\cite{inpars}.

\textbf{Neural Readers.}
The reader constitutes the other fundamental component of a contemporary QA system, typically implemented as a neural machine reading comprehension (MRC) model. 
Recently, most QA systems have introduced Generative Readers into their architecture~\cite{lewis2020retrieval}.
RFiD~\cite{wang2023rfid} introduces a relational fusion-in-decoder (FiD) model. By concurrently referencing multiple passages, RFiD utilizes the encoders of FiD to distinguish between causal relations and spurious features, effectively guiding the generation process. More recently, the remarkable capabilities exhibited by LLMs in comprehending\cite{peng2023large}, extracting\cite{xu2023large}, and reasoning\cite{peng2023gpt,wu2023survey} with intricate information have prompted the adoption of larger LLMs, such as GPT-3 and ChatGPT, for answer generation. For example, In-Context RALM~\cite{ram2023context} is an in-context retrieval-augmented language modeling method, achieving off-the-shelf general-purpose retrievers without the need for further language model training.

\section{Preliminaries}

\begin{table}[t]
    \centering
    \label{tab:qa_example}
    \caption{Examples of various types of QA pairs. The sensitive information has been desensitized and is represented by ``xxx".}
    
    \setlength{\tabcolsep}{1mm} 
    \begin{tabular}{p{1.5cm} p{4cm} p{6cm} }
    \toprule
    Types & Questions & Answers  \\ \hline
    Fact-oriented & When did the xxx operational strategy launch? & It launched on March 21, 2021.  \\ \hline
    Solution-oriented (Short) & How can the strategic depth of RTS (Real-Time Strategy) games in xxx be increased? & It can be enhanced by adding positional movement and control operations for each card during combat. \\ \hline
    Solution-oriented (Long) & How does xxx recommendation system update in a timely manner to meet user preferences when user interests change?  & Step One: Identify changes in user interests through recording behavior data and using data analysis. .... Step Two: Update and optimize the recommendation model in real time... Step Three: Evaluate the feedback after the system update for continuous optimization... \\ \bottomrule
    \end{tabular}

\end{table}

In this section, we describe an enterprise knowledge base from a technology company, present the process of developing a test set for enterprise knowledge question answering, and formally define our research problem.
\subsection{Enterprise Knowledge Base}

We have gathered an enterprise knowledge base from a high-tech company mainly consisting of technical and project documentation. The ongoing expansion of these documents, produced by the company's workforce, emphasizes the need for an effective knowledge management system. We filtered out documents shorter than 50 words, as they typically lack comprehensive information, resulting in an enterprise knowledge base of 291,649 documents.


\subsection{Construction of QA Test Sets}

In real-world industrial scenarios, there is a notable lack of accumulated QA data for enterprise knowledge bases. In response, 
we categorized potential questions from the collected knowledge base into types: ``Who/Whose/Whom'', ``Where'', ``When'', ``What'', ``Which'', and ``How''. We classified all types except ``How'' as fact-oriented questions while categorizing ``How''-type as solution-oriented questions. 

Table 1 showcases examples of both fact-oriented and solution-oriented questions. We collaborated with domain experts to pose questions and provide answers for a randomly chosen collection of 500 documents from the enterprise knowledge base, culminating in the creation of a high-quality testing dataset. Notably, we further categorized the solution-oriented test set based on the answer length in solution-oriented QA pairs. Longer solution-oriented QA pairs typically involve queries and responses about solutions to complex scenarios. Detailed data statistics can be found in Table 2. In the experimental section, we will describe how to utilize these test sets to validate our system.

\begin{table}[t]
    \centering
    \label{tab:EK datasets}
    \caption{Statistics of QA test set for enterprise knowledge base.}
    \begin{tabular}{cccc}
    \toprule
    Types  & Fact-oriented  & Solution-oriented (Short) & Solution-oriented (Long)  \\
    \midrule 
      Number      &  1,386   & 167    & 234  \\
      Query Length & 24.40  & 30.00  & 48.24  \\
      Answer Length & 27.08  & 58.94 & 298.14  \\
      \bottomrule
    \end{tabular}
\end{table}

\subsection{Problem Definition}

\begin{definition}

\textbf{Question Answering for the Enterprise Knowledge Base.}
Given a question $q$ and an enterprise knowledge base $\mathcal{D}=\left\{d_1, \ldots, d_{|\mathcal{D}|}\right\}$, where $|\mathcal{D}|$ is the size of documents in the enterprise knowledge base $\mathcal{D}$, our goal is to develop a system $\mathcal{S}$, which can generate precise and coherent answer to the question $q$, utilizing the information contained within $\mathcal{D}$.
\end{definition}


\section{Method}
\begin{figure}[t]
\centering
\includegraphics[width=0.75\linewidth]{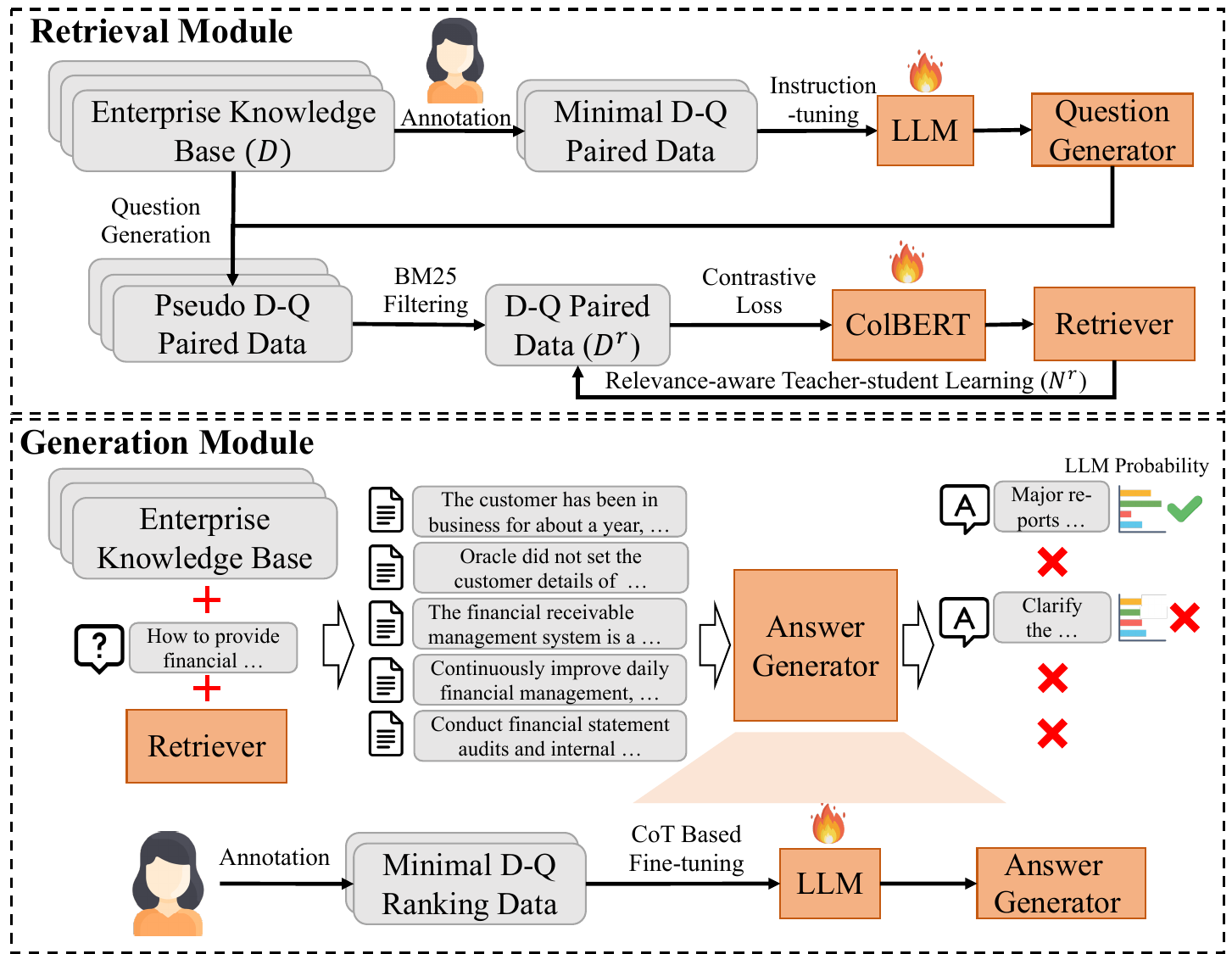} 
\caption{The illustration of EKRG framework.}
\label{fig:prompt}
\end{figure}

In this section, we introduce the details of our EKRG framework. As illustrated in Figure~\ref{fig:prompt}, it encompasses two primary modules: a retrieval module that sources relevant documents from $\mathcal{D}$ in response to user queries, and a generation module that crafts precise and coherent answers derived from these retrieved documents.


\subsection{Retrieval Module}
To efficiently retrieve documents relevant to user queries, we have constructed our retrieval module utilizing the ColBERT~\cite{khattab2020colbert} architecture. Specifically, for a given query $q_i$ and document $d_j$, ColBERT computes their embedding $E_{q_i}\in \mathbf{R}^{|q_i|\times dim}$ and $E_{d_j}\in \mathbf{R}^{|d_j|\times dim}$. Subsequently, the relevance score $S_{q_i,d_j}$ is calculated as follows,  
\begin{equation}
    S_{q_i, d_j} = \sum_{m\in[1, |q_i|]} \max_{n\in [1,|d_j|]} \frac{E_{q_{i,m}} \cdot E_{d_{j,n}}}{\|E_{q_{i,m}}\| \| E_{d_{j,n}}\|},
\end{equation}
where $|q_i|$ and $|d_j|$ denote the lengths of $q_i$ and $d_j$ respectively, $E_{q_{i,m}}$ and $E_{d_{j,n}}$ is the $m$-th token vector of $q_i$ and $n$-th token vector of $d_j$, $\|\cdot\|$ denotes the vector length and $\cdot$ symbolizes the dot product operation.

However, as we mentioned above, enterprise knowledge bases often lack annotated data, i.e., document-question pairs, for training the dense retriever. To overcome this challenge, we innovate a query generation model based on the LLM, specifically designed to construct document-question pairs for training ColBERT. Additionally, we introduce a relevance-aware teacher-student learning strategy, significantly improving the training process's efficiency.

\subsubsection{Pseudo Document-Question Pairs Generation} 

Recently, LLMs have been extensively validated for their generative capabilities in the few-shot learning scenarios. Intuitively, we can adapt a method akin to Inpars~\cite{inpars}, which utilizes the GPT-3 API to generate potential questions for assorted documents. However, given the stringent privacy and security protocols that typically regulate enterprise knowledge bases, we utilize instruction-tuning to train the deployable, private LLMs. In this scenario, dealing with non-English corpus, we specifically integrate an open-source LLM, i.e., Baichuan-13B as the backbone. 

For training, we utilize a minimal amount of document-query pairs annotated by experts. And to address the two types of questions previously introduced, i.e., fact-oriented and solution-oriented, we develop two distinct instructions as follows.

\begin{tcolorbox}[colback=white,colframe=black!50!white,arc=2mm,boxrule=1pt,boxsep=0pt,left=6pt,right=6pt,top=6pt,bottom=6pt,boxrule=1pt,title={Instruction for generating fact-oriented questions},fonttitle=\bfseries]
\texttt{\small Please carefully review the provided document and formulate specific questions that directly relate to its content. It is important that the answers to these questions can be located within the document itself. }
\end{tcolorbox}

\begin{tcolorbox}[colback=white,colframe=black!50!white,arc=2mm,boxrule=1pt,boxsep=0pt,left=6pt,right=6pt,top=6pt,bottom=6pt,boxrule=1pt,title={Instruction for generating solution-oriented questions},fonttitle=\bfseries]
\texttt{\small Please conduct a thorough analysis of the project or technical background outlined in the document. Begin by summarizing the current situation, the specific task at hand, the proposed solution, and the eventual outcome. Additionally, formulate specific questions that effectively integrate the given scenarios and tasks. }
\end{tcolorbox}

Additionally, fine-tuning the entire LLM model requires significant computational resources and time. To overcome this challenge, we utilize a parameter-efficient fine-tuning approach called LoRA~\cite{hu2022lora}. This method entails fixing the parameters of the pre-trained LLM and training rank decomposition matrices specific to each layer in the Transformer architecture. The learning objective is formulated as follows,



\begin{equation}
 \max_{{\Theta^L}^{\prime}} \sum_{(x, y) \in \mathcal{T}} \sum_{t=1}^{|y|} \log \left(P_{{\Theta^L}+{\Theta^L}^{\prime}}\left(y_t \mid x, y_{<t}\right)\right),
\end{equation}
where  $\Theta^L$ is the original parameters of LLM, ${\Theta^L}^{\prime}$ is the LoRA parameters which will be updated during the training process, $x$ and $y$ represent the ``Instruction Input" and ``Instruction Output" in the training set $\mathcal{T}$, $y_t$ is the $t-th$ token of $y$, $y_{<t}$ represents the tokens before $y_t$.


\subsubsection{Relevance-aware Teacher-Student Learning Strategy}

Given the inconsistent quality of pseudo-question production by LLM, subpar samples can negatively impact the training efficacy of dense retrievers. In response, we devise a relevance-aware teacher-student learning strategy to optimize training efficiency and effectiveness. Motivation is to dynamically select high-quality queries when training the retriever model.  

We first introduce a strategy for query selection, whereby a high-quality generated query should reliably retrieve its source document, a process referred to as consistency filtering.Consistency filtering has been proven effective in generating synthetic questions for QA tasks\cite{lewis2021paq}. Specifically, for a generated document-query pair $\left(q^*,d\right)$, a retriever is employed to identify the top-k documents most relevant to $q^*$ within a corpus. If $d$ is within this top-k selection, the synthetic query $q^*$
is retained for subsequent training steps. To train our dense retriever model, we initially utilize BM25 to curate a small subset of high-quality queries as the foundational training data through consistency filtering. This initial step is crucial as the baseline dense retriever model may underperform at the onset, whereas BM25 tends to offer stable performance.

Following this, we utilize the selected data for an initial training phase, often referred to as the warm-up step of the retrieval model. After this preliminary training, the dense retrieval model demonstrates a significant sensitivity to the corpus. Subsequently, we engaged in iterative training of the dense retrieval model and the training data is selected by models from last step using consistency filtering strategy. During this phase, queries that are not well recalled are selected for the next training cycle. Additionally, we exclude examples with particularly poor recall performance to avoid introducing noise from the query generation process. The detailed algorithm is outlined in Algorithm~\ref{alg:1}. Through this teacher-student learning strategy, we substantially minimize the number of training instances required.




\begin{algorithm}[t]
\small
\renewcommand{\algorithmicrequire}{\textbf{Input:}}
\renewcommand{\algorithmicensure}{\textbf{Output:}}
\caption{Relevance-aware teacher-student learning strategy}
\label{alg:1}
\begin{algorithmic}[1]
    
\REQUIRE Given the generated query-document pairs set $Q_g = \{\left(q_i,d_j \right) \}$, the number of iteration steps $T$, a lexical retrieval method BM25 and a initial retrieval model $R_0$.
\ENSURE The final retrieval model $R_{T}$. 
\STATE { $D^r \leftarrow$ use BM25 to select 1k seed q-d pairs from $Q_g$}
\STATE { $R_1 \leftarrow$ update retrieval model $R_0$ with $D^r$ by Eq. \ref{alg:infoECE}}
\FOR{iteration step $t$ to $T$}
\STATE { $N^r \leftarrow$ use $R_{t-1}$ to select q-d pairs from $Q_g$ }
\STATE {  $D^r \leftarrow$ $N^r \cup D^r$}
\STATE { $R_t \leftarrow$ update retrieval model $R_{t-1}$ with $D^r$ by Eq.\ref{alg:infoECE}}
\ENDFOR
\end{algorithmic}  
\end{algorithm}

Recent studies have shown that negative sampling techniques can improve the effectiveness of representation learning\cite{hu2023boss, wang2020setrank,luo2024survey,qin2019duerquiz}.
To further enhance the performance of retrieval, in each training round, for every generated document-query pair $(q,d^+)$, we randomly select a set of irrelevant documents $\{{d_1}^-,\ldots,{d_k}^-\}$ that is retrieved by BM25 in top-1000, as the negative pairs. Then, we minimize the InfoNCE loss to train the retriever as follows:

\begin{equation}
\label{alg:infoECE}
\mathcal{L}(\theta) = -\log \frac{\exp \left(S_{q, d^+}\right)}{\exp \left(S_{q, d^+}\right)+\sum_{l=1}^k \exp \left(S_{q, {d_l}^-}\right)} .
\end{equation}


\subsection{Generation Module}
In the generation phase, we aim to craft precise and coherent answers based on the top-$k$ retrieved documents in response to each user query. Here, capitalizing on the generative prowess of LLMs, we introduce a novel CoT-based fine-tuning method for answer generation.

\subsubsection{CoT-based Fine-tuning}
By leveraging a small set of annotated question-answer pairs, we develop an instruction-turning dataset for fine-tuning an LLM, enabling it to answer questions based on retrieved documents. Formally, we denote the annotated question-answer pairs as $\mathcal{D}^G= \{\left( q^i,y^i \right)| i \in [1, |\mathcal{D}^G]|\}$, where $|\mathcal{D}^G|$ represents its size. For each question-answer pair, we employ the above retrieval module to obtain the top-$k$ associated documents $\{d^i_1,\ldots,d^i_k \}$. Along this line, we create a separate fine-tuning example for each relevant document by adding it to the instructions as input, i.e., $\left\{\left(d^i_j \circ q^i, y^i\right) \mid j=1 \ldots k \right\}$. This means that for each QA pair, we generate $k$ unique training instance. Intentionally, we refrain from combining the top-$k$ retrieved documents into one input to minimize input length. It helps prevent issues with LLMs struggling to generate accurate responses when confronted with overly lengthy inputs. Correspondingly, we will subsequently discuss how to integrate answers based on different retrieved documents given a specific question. By employing LoRA to enhance the training efficiency, the learning objective for the generator is defined as follows:



\begin{equation}
 \max_{{\Theta^G}^{\prime}} \sum_{(q^i, y^i) \in \mathcal{D}^G} \sum_{j=1}^{k} \sum_{t=1}^{|y^i|} \log \left(P_{\Theta^G+{\Theta^G}^{\prime}}\left(y^i_t \mid d^i_j \circ q^i, y^i_{<t}\right)\right),
\end{equation}
where  $\Theta^G$ is the parameters of generator $G$ and  ${\Theta^G}^{\prime}$ is the LoRA parameters.

Considering that retrieved documents may not always answer the user's question, we expect LLMs to discern whether a given document contains the information to answer the question. Inspired by the CoT~\cite{wei2022chain}, we break down this instruction into several steps. Initially, the model should summarize the provided document for a comprehensive understanding. Then, it assesses whether the document directly addresses the question. If so, the model generates a final response based on the summarized information. Otherwise, if the document is deemed irrelevant, the model issues the response as ``irrelevant''. Additionally, our proposed CoT fine-tuning method effectively mitigates hallucinations in LLMs, enabling them to answer questions based on the provided knowledge documents. The instruction for CoT fine-tuning is as follows, 


\begin{tcolorbox}[colback=white,colframe=black!50!white,arc=2mm,boxrule=1pt,boxsep=0pt,left=6pt,right=6pt,top=6pt,bottom=6pt,boxrule=1pt,title={Instruction for CoT-based fine-tuning},fonttitle=\bfseries]
\texttt{\small Please follow the steps below to ensure an accurate response: \\ 
1. Thoroughly read the provided document and provide a brief yet comprehensive summary of its main points. \\
2. Assess and elucidate how the given document can be applied to address the given question. \\ 
3. If the document is deemed irrelevant, indicate this by outputting the term "irrelevant". Otherwise, provide the answer along with the corresponding supporting information.}
\end{tcolorbox}

\subsubsection{Answers Integration} To aggregate the $k$ answers based on the various retrieved documents into a final conclusion, we calculate the language model probability $p_y$ of the answer $y$ given the input context $x$ and a query $q$ as follows:

\begin{equation}
p_y=\frac{1}{|y|} \sum_{i=1}^{|y|} \log p\left(y_i \mid x, q, y_{<i}\right).
\end{equation}

Next, we choose the answer with the highest language model probability, excluding any responses deemed ``irrelevant''. Moreover, if the answer is not among the initial top-$k$ retrieved passages, we continue iterating through the retrieval process until the answer is found or until we reach the predefined maximum number of retrieval steps.


\section{Experiments}


\subsection{Experiment Settings}



\subsubsection{Datasets}
We conducted a thorough assessment of our model's efficacy on both retrieval and generation. In the field of retrieval, we created our retrieval dataset, named EKR, by utilizing the Enterprise Knowledge base. For each question, we treat the corresponding document as its retrieval ground truth. To further enhance the evaluation, we integrated three supplementary public datasets: FiQA, NQ, and MS MARCO, achieving a holistic validation of our model's adeptness in general retrieval. On the generation aspect, we validate the generative ability on three subcategory QA datasets , explicitly showcasing our model's proficiency across diverse contextual scenarios.


To train our question generation LLMs and the question answering LLM, we annotated a small amount of training data. Specifically, for the task of generating pseudo questions using instruction tuning, we labeled 50 document-question pairs, encompassing the three types of questions. For the CoT-based question answering task, we annotated 50 question-answer pairs using the CoT method, where each question involved retrieving the top-5 relevant documents, resulting in a total of 250 answers. 

\subsubsection{Evaluation Metrics}
In the context of information retrieval, we employed four widely recognized evaluation metrics NDCG@10, MAP@10, Recall@10, and MRR@10 to gauge the improvement of retrieval models facilitated by synthesized questions. 
For factual questions, we utilized EM (Exact Match)  and F1 score as metrics to evaluate the consistency achieved by our model in generating accurate responses to factual inquiries. In the realm of solution-oriented questions, we employed BLEU and ROUGE  metrics to appraise the fluency and accuracy of the generated answers. 

\subsubsection{Implementation Details}
We engaged in the fine-tuning process with two prominent open-source LLMs, namely Baichuan-13b \cite{zeng2023evaluating} and LLaMA-7b \cite{touvron2023llama}.
For our enterprise knowledge base, we employed Baichuan-13b, pre-trained on the Chinese corpus, as the foundational backbone. Conversely, on the English datasets FiQA, NQ, and MS MARCO, we leveraged LLaMA-7b as the backbone, specifically tailored to the English corpus.

All experiments were conducted on a high-performance setup featuring 8 Tesla A800 80G GPUs. The fine-tuning process for all LLMs employed the LoRA method with a LoRA-rank set to 8, and model parameters were optimized using the Adam optimizer with a default learning rate of 5e-5.
In our methodology, ColBERT served as the dense retriever, configured with a batch size of 32 and a maximum text length of 350, complemented by a learning rate of 2e-5. 

\subsection{Baseline Methods}


In the realm of retrieval, we compared various unsupervised retrieval baselines, encompassing both traditional data augmentation methods and contemporary approaches centered on LLMs:
\begin{itemize} 
    \item $\textbf{ICT}$~\cite{lee2019latent} randomly selects a sentence from the provided document as a query to train the retriever.
    \item $\textbf{Contriever}$~\cite{izacard2021unsupervised} constructs a pseudo query-document by randomly extracting two spans from the document for training an unsupervised dense retriever.
    \item $\textbf{BM25}$~\cite{robertson1995okapi} is an evolution of the earlier TF-IDF (Term Frequency-Inverse Document Frequency) algorithm, widely adopted as a ranking function in information retrieval systems.
    \item $\textbf{Inpars}$~\cite{inpars} leverages human-written prompts to guide close-source LLMs, such as Curie from OpenAI, in generating queries.
    \item $\textbf{SPTAR}$~\cite{peng2023soft} employs an optimized soft prompt to guide open-source LLMs, like LLaMA, in generating weak document-query pairs.
\end{itemize}
We also conducted comparisons with several most recent open-domain question answering and retrieval-augmented generation baselines:
\begin{itemize} 
    \item $\textbf{RAG}$~\cite{lewis2020retrieval} constitutes an end-to-end model comprising a pre-trained neural retriever and a pre-trained seq2seq transformer. It explores a general-purpose fine-tuning recipe for retrieval-augmented generation.
    \item $\textbf{RFiD}$~\cite{wang2023rfid} introduces a relational Fusion-in-Decoder model by concurrently referencing multiple passages. 
    \item $\textbf{In-Context RALM}$~\cite{ram2023context} is an In-Context Retrieval-Augmented Language Modeling method, achieving off-the-shelf general-purpose retrievers without the need for further LM training.
\end{itemize}


\subsection{Experimental Results}

\subsubsection{Retrieval}
\begin{table}[t]
\centering
\caption{Performance comparisons on retrieval compared with baselines.}
\begin{tabular}{p{2cm}|p{1.8cm}|>{\centering\arraybackslash}p{1.2cm}>{\centering\arraybackslash}p{1.2cm}>{\centering\arraybackslash}p{1.2cm}>{\centering\arraybackslash}p{1.2cm}>{\centering\arraybackslash}p{1.2cm}>{\centering\arraybackslash}p{1.2cm}}
\toprule
Datasets &Metrics & BM25 & Contriver & ICT & Inpars & SPTAR & Our \\ \hline
\multirow{4}{*}{EKR} & NDCG@10 & 0.6934 & 0.7161 & 0.6914 & 0.8078 & 0.8147 & 0.8723 \\ 
&MAP@10       & 0.6425 & 0.6591 & 0.6287 & 0.7749 & 0.7814 & 0.8265\\
 & Recall@10 & 0.7834 & 0.7885 & 0.7655 & 0.8735 & 0.8786 & 0.9189 \\ 
 &MRR@10 & 0.6647 & 0.6932 & 0.6678 & 0.7867 & 0.7942 & 0.8560\\ \hline
\multirow{4}{*}{FiQA}& NDCG@10 & 0.2361 & 0.2536 & 0.1955 & 0.2574 & 0.2688 & 0.2745 \\ 
&MAP@10 & 0.1784 & 0.2002 & 0.1515 & 0.2024 & 0.2103 & 0.2124\\
& Recall@10 & 0.2951 & 0.2994 & 0.2278 & 0.3051 & 0.3083 & 0.3402 \\ 
&MRR@10 & 0.2889 & 0.3259 & 0.2585 & 0.3179 & 0.3039 & 0.3435\\ \hline
\multirow{4}{*}{NQ}& NDCG@10 & 0.2855 & 0.2538 & 0.2601 & 0.3291 & 0.3307 & 0.3685 \\ 
&MAP@10 & 0.2454 & 0.197 & 0.1917 & 0.3012 & 0.2697 & 0.2641\\
 &Recall@10 & 0.4555 & 0.4128 & 0.4012 & 0.4341 & 0.4953 & 0.5053 \\ 
 &MRR@10 & 0.2634 & 0.2155 & 0.2077 & 0.3373 & 0.2807 & 0.3064 \\\hline
\multirow{4}{*}{MS MARCO}& NDCG@10 & 0.2284 & 0.2056 & 0.1389 & 0.1821 & 0.2285 & 0.2344 \\ 
 &MAP@10 & 0.1803 & 0.1578 & 0.1095 & 0.1444 & 0.1872 & 0.1899\\
  &Recall@10 & 0.3787 & 0.3568 & 0.2112 & 0.2991 & 0.3662 & 0.3707 \\
 &MRR@10 & 0.1733 & 0.1611 & 0.098 & 0.148 & 0.1704 & 0.2034 \\ 
 \bottomrule
\end{tabular}
\label{tab:retrieve result}
\end{table}

\begin{table}[t]
\centering
\caption{Performance comparisons on generation compared with baselines.}
\begin{tabular}{p{2.7cm}|>{\centering\arraybackslash}p{0.8cm}>{\centering\arraybackslash}p{0.8cm}|>{\centering\arraybackslash}p{1.7cm}>{\centering\arraybackslash}p{1.7cm}|>{\centering\arraybackslash}p{1.7cm}>{\centering\arraybackslash}p{1.7cm}}
\toprule
  
& \multicolumn{2}{c|}{Fact-oriented} & \multicolumn{2}{c|}{Solution-oriented (Long)} & \multicolumn{2}{c}{Solution-oriented (Short)} \\
Metric & EM & F1 & BLEU & ROUGE & BLEU & ROUGE \\ \hline
RAG   & 31.76 & 30.10 & 18.54 & 24.01 &30.03 & 39.28  \\
RFiD & 42.34 & 39.28 & 20.60 & 31.98 &36.70 & 41.95 \\
In-Context RALM & 45.44 & 44.41 & 33.59 & 50.51 &47.58 & 64.60 \\
EKRG & 55.74 & 48.29 & 45.68 & 68.39 & 57.96 & 74.86 \\ \hline
- w/o CoT & 51.09 & 47.14 & 44.64 & 67.77 & 50.14 & 68.16  \\
\bottomrule
\end{tabular}
\label{tab:kg result}
\end{table}



As shown in Table~\ref{tab:retrieve result}, traditional data augmentation methods such as ICT fall short of the performance of the frequency statistics-based BM25 across various datasets. In contrast to traditional data augmentation methods, the utilization of in-context learning based on LLMs for generating pseudo questions aligns more closely with the authentic distribution of questions. As evident in Table~\ref{tab:retrieve result}, Inpars outperforms BM25 in terms of retrieval on EKR, FiQA, and NQ datasets. 
Moreover, when compared to Inpars, SPTAR elevates the quality of generated queries through soft prompt learning, thereby demonstrating the effectiveness of training on LLMs.

Our proposed method exhibits superior performance compared to all baselines, with the most substantial gains observed on FiQA and EKR relative to SPTAR. This underscores that, through diversity-driven instruction tuning, our model excels at generating questions that better conform to the distribution of document types, even with minimal reliance on labeled data.


\subsubsection{Generation}
As shown in the Table \ref{tab:kg result}, the performance of generation based on LLMs exceeds that of traditional generative models like RAG and RFiD, suggesting that extensive pre-training on a vast corpus significantly enhances LLMs' text comprehension abilities. Especially in situations with limited training data, traditional models are insufficient to learn enough useful information, resulting in poorer summarization and analytical abilities. However, with the use of LLMs, merely utilizing the capability of in-context learning, RALM's performance surpasses that of traditional generative models after training.

Our EKRG approach outperforms RALM by effectively fine-tuning LLMs to answer questions accurately using retrieved texts, even with limited training data, as indicated in Table \ref{tab:kg result} under EKRG \textsl{\mbox{-w/o CoT}}. Moreover, due to the reasoning abilities of the CoT method, EKRG more effectively discerns and excludes irrelevant documents, while providing clear explanations for the utilized ones. This approach enhances performance beyond mere fine-tuning, as demonstrated in Table \ref{tab:kg result}.

\subsection{Ablation Study}
\begin{figure}[t]
    \centering
    \includegraphics[width=\linewidth]{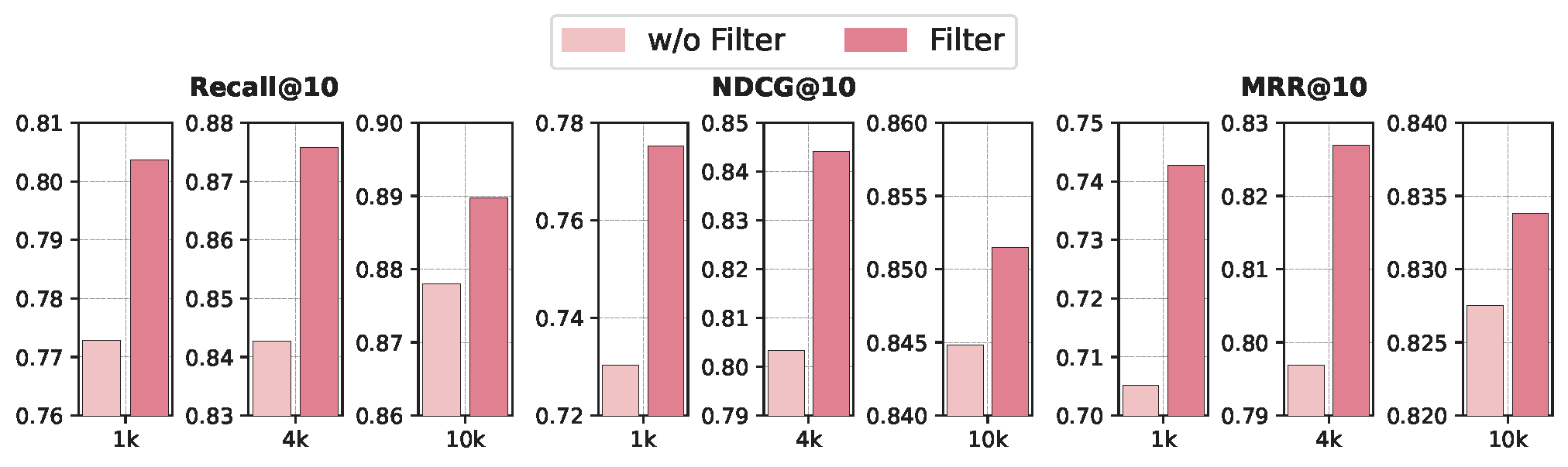}
    \caption{The results of ablation study on the high-quality query filtering.}
    \label{fig:abalation}
\end{figure}

\textbf{The ablation study on the high-quality query filtering.}
In this section, we present the noteworthy impacts of high-quality query filtering. As depicted in Figure \ref{fig:abalation}, queries subjected to our filtering method consistently outperform randomly selected queries when utilizing an equivalent number of training queries. Additionally, our filtering approach enables achieving comparable results to using the entire query set with significantly fewer queries. This underscores the efficiency and effectiveness of our high-quality query filtering method.

\textbf{The ablation study on CoT based fine-tuning.}
In this segment, we showcase the effectiveness of CoT-based fine-tuning algorithm. As indicated in Table \ref{tab:kg result}, CoT-based fine-tuning improves the performance of Solution-oriented (Long)-type question answering. This enhancement suggests that the model more accurately identifies truly relevant documents within the top-k retrieved, filtering out irrelevant ones despite their high retrieval relevance scores.


\subsection{Case Study}

As illustrated in Table \ref{tab:case}, we demonstrated the responses generated by EKRG to highlight the benefits of using CoT. EKRG determined that Document \#1 $\left(D \#1\right)$ aimed to enhance the quality of nighttime forest images and did not propose a real-time image dehazing algorithm, thus providing an ``irrelevant'' response. EKRG considered that D\#2 used software to implement image dehazing functions, which was somewhat related, so it summarized the answer based on D\#2. EKRG believed that D\#3 mentioned a real-time dehazing method, which was most relevant to the question, and provided the final answer. EKRG selected the third response as the definitive answer, which possesses the highest likelihood according to the language model probability.

\begin{table*}[t]
\small
\caption{Case study on the generation results.}
\begin{tabularx}{\textwidth}{X}
\toprule
\textbf{Q}: How to use real-time image defogging technology to solve the blur problem of images collected in foggy days?\\
\hline

\textbf{D \#1}: Aiming at the problems of low contrast and high noise in nighttime forest images, this research is dedicated to ... Technical research: Conduct in-depth research on various nighttime image enhancement methods, including image fusion methods, dehazing model methods and machine learning methods... \\

\textbf{D \#2}: Many features in images obtained in foggy environments are blurred, and the function of machine vision systems based on image features will be affected...Based on the dark channel prior defogging algorithm, the image is dehazed through Matlab software, the system parameters are adjusted to find the optimal solution for dehazing...\\

\textbf{D \#3}: If the image equipment collects images in foggy days, the collected images will have problems such as low brightness and blurred image outlines... developed using Xilinx FPGA, the core processor is Artix-7;  Image processing module: ... At the hardware level, the shift register implements the filtering function... \\
\hline

\textbf{A \#1}: This document examines ways to improve the quality of nighttime forest images. It does not introduce the specific defogging methods, so it is not relevant to the problem. Answer: Irrelevant.\\

\textbf{A \#2}: This document introduces how to eliminate haze on image imaging. This document proposes some relevant methods to the question. Answer: The first step is to improve the dark channel prior defogging algorithm. In the second step, the image is dehazed through matlab software. The third step is to...  (language model probability: 0.8687)  \\

\textbf{A \#3}: This document mainly solves the problem of blurred outlines in images collected in foggy days, and proposes solutions in terms of algorithms and hardware. This article provides specific solutions. Answer: The first step is to develop systems using Xilinx with the core processor Artix-7. The second step is to use the physical model of atmospheric light transmission and the dark channel method to perform real-time dehazing and restoration processing of hazy images. In the third step... (language model probability: 0.9718) \\
\bottomrule
\end{tabularx}
\label{tab:case}
\end{table*}

\section{Conclusion}
In this work, we proposed a novel framework, termed EKRG, which is based on LLM to enable question-answering for enterprise knowledge bases with limited annotation costs. We first introduced an instruction-tuning method using LLM to enhance the training of knowledge retriever. Then we developed a relevance-aware teacher-student learning strategy to further enhance the efficiency of the training process. Additionally, we proposed a novel CoT based fine-tuning method to empower the LLM-based knowledge generator. Finally, extensive experiments on real-world datasets have demonstrated the effectiveness of our proposed framework.

\noindent\textbf{Acknowledgements.} This work was supported in part by Guangzhou-HKUST Joint Funding Program(Grant No.2023A03J0008), the National Natural Science Foundation of China under Grant No. 62176014, the Fundamental Research Funds for the Central Universities, Education Bureau of Guangzhou Municipality, Guangdong Science and Technology Department, Foshan HKUST Projects(FSUST21-FYTRI01A), Postdoctoral Fellowship Program of CPSF under Grant Number GZC20232811. 

\bibliographystyle{splncs04}

\bibliography{reference}

\end{document}